\pdfoutput=1

\documentclass[11pt]{article}

\usepackage[]{acl}

\usepackage{times}
\usepackage{latexsym}
\usepackage{graphicx}
\usepackage{booktabs}
\usepackage{multirow}
\usepackage{comment} 
\usepackage{tabularx}
\usepackage{xcolor}
\usepackage[T1]{fontenc}

\usepackage[utf8]{inputenc}

\usepackage{microtype}

\newcommand{\orig}{\textsc{Orig}}
\newcommand{\nopos}{\textsc{Nopos}}

\newcommand{\sn}{\textsc{Shuf.n1}}
\newcommand{\snn}{\textsc{Shuf.n2}}
\newcommand{\snnn}{\textsc{Shuf.n3}}
\newcommand{\snnnn}{\textsc{Shuf.n4}}
\newcommand{\scorp}{\textsc{Shuf.Corpus}}

\title{Word Order Does Matter \\\textit{(And Shuffled Language Models Know It)}}

\author{\thanks{~~Equal contribution. Order was decided by a coin toss.}\, Vinit Ravishankar\footnotemark[2]\quad \footnotemark[1]\, Mostafa Abdou\footnotemark[3]\quad
Artur Kulmizev\footnotemark[4]\quad Anders Søgaard\footnotemark[3]\vspace{0.3em} \\
  \footnotemark[2]\, Language Technology Group, Department of Informatics, University of Oslo \\
  \footnotemark[3]\, Department of Computer Science, University of Copenhagen \\
  \footnotemark[4]\, Department of Linguistics and Philology, Uppsala University\vspace{0.3em} \\
  \footnotemark[2]{~~\tt vinitr@ifi.uio.no} \\
  \footnotemark[3]{~~\tt \{abdou,soegaard\}@di.ku.dk}\\
}

\date{}

\begin{document}
\maketitle
\begin{abstract}

Recent studies have shown that language models pretrained and/or fine-tuned on randomly permuted sentences exhibit competitive performance on GLUE, putting into question the importance of word order information. Somewhat counter-intuitively, some of these studies also report that position embeddings appear to be crucial for models' good performance with shuffled text. We probe these language models for word order information and investigate what position embeddings learned from shuffled text encode, showing that these models retain information pertaining to the original, naturalistic word order. We show this is in part due to a subtlety in how shuffling is implemented in previous work -- {\em before} rather than {\em after} subword segmentation. Surprisingly, we find even Language models trained on text shuffled {\em after} subword segmentation retain some semblance of information about word order because of the statistical dependencies between sentence length and unigram probabilities. Finally, we show that beyond GLUE, a variety of language understanding tasks do require word order information, often to an extent that cannot be learned through fine-tuning.

\end{abstract}
\section{Introduction}
Transformers \cite{vaswani2017attention}, when used in the context of masked language modelling \cite{devlin2018bert}, consume their inputs concurrently. There is no notion of inherent order, unlike in autoregressive setups, where the input is consumed token by token. To compensate for this absence of linear order, the transformer architecture originally proposed in~\citet{vaswani2017attention} includes a fixed, sinusoidal position embedding added to each token embedding; each token carries a different position embedding, corresponding to its position in the sentence. The transformer-based BERT~\citep{devlin2018bert} replaces these fixed sinusoidal embeddings with unique, learned embeddings per position; RoBERTa~\citep{LiuRoBERTaRobustlyOptimized19a}, the model investigated in this work, does the same.

\begin{figure}
    \centering
    \includegraphics[width=2.9in]{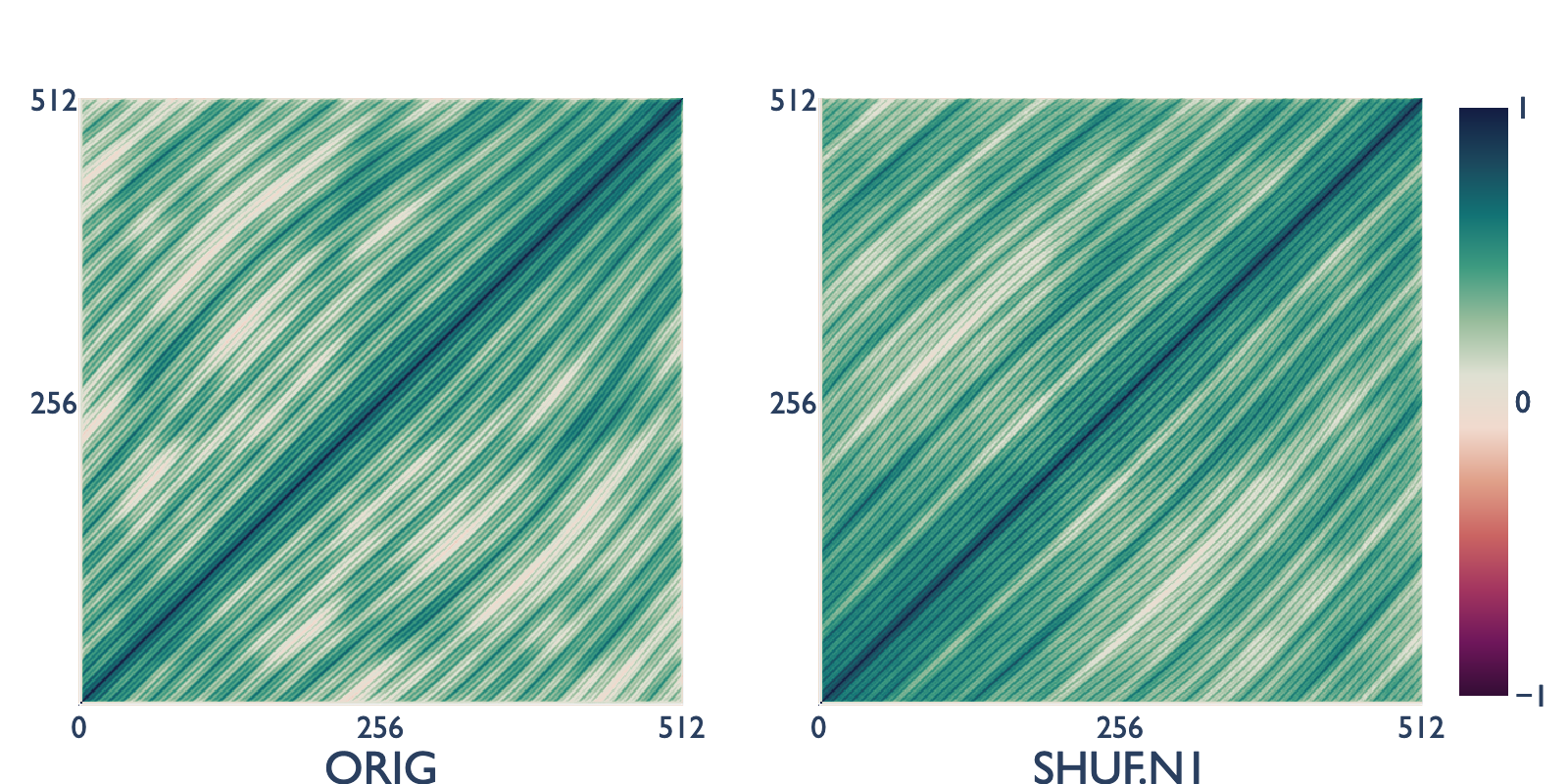}
    \caption{Pearson correlations between position embeddings for full-scale models; the patterns are similar to fully learnable absolute embeddings~\citep{wang_positionembeddings_2021} and can be said to have learned something about position. We later demonstrate that this is not the case with post-BPE scrambling.}
    \label{fig:pos_dot}
\end{figure}

\begin{figure*}
    \centering
    \includegraphics[width=\textwidth]{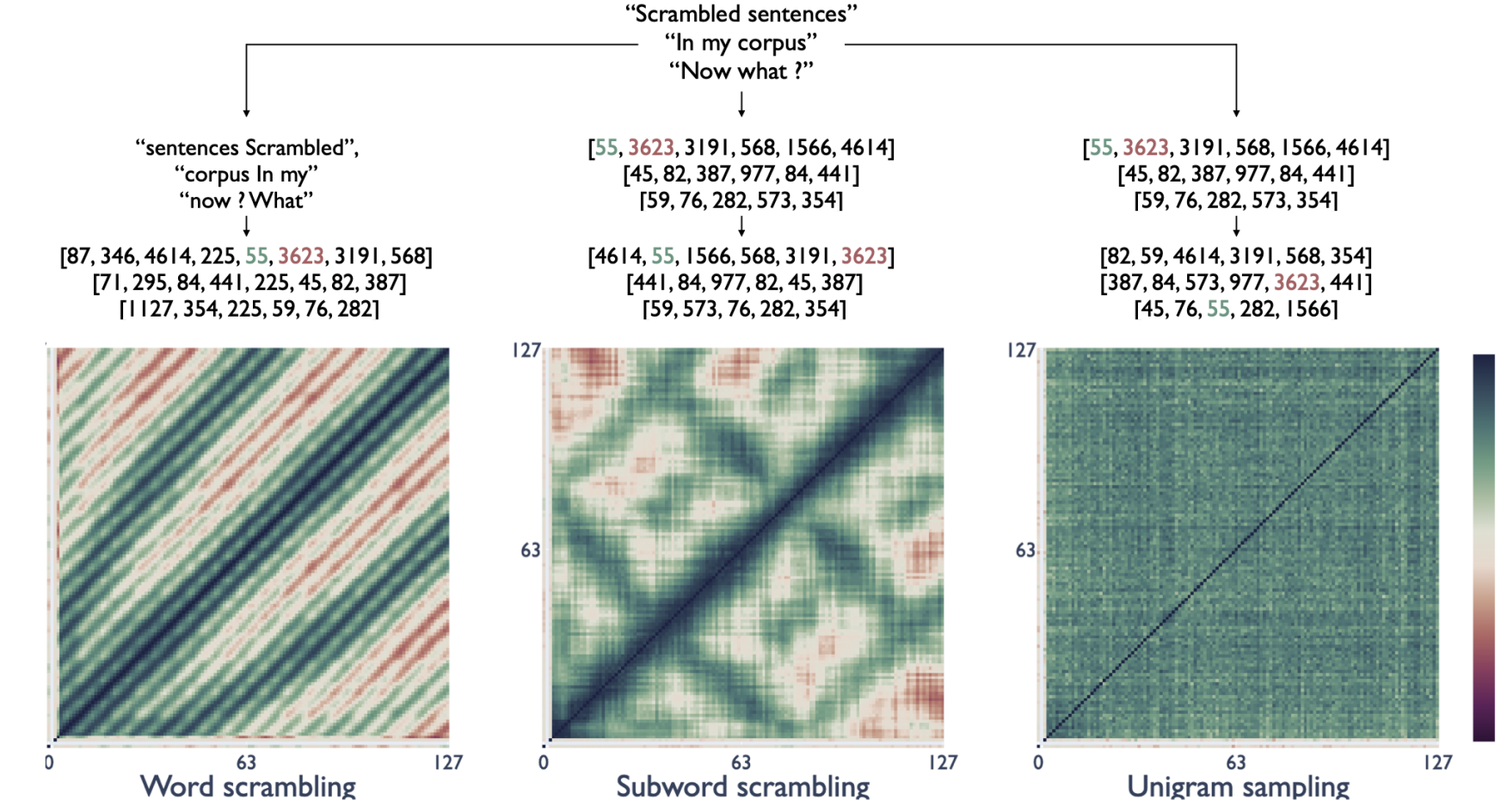}
    \caption{Correlations between position embeddings when shuffling training data {\em before} segmentation (left), i.e, at the word level, and {\em after} segmentation (middle), i.e., at the subword level, as well as when replacing all subwords with random subwords based on their corpus-level frequencies (right). The latter removes any dependency between subword probability and sentence length. The plots show that shuffling before segmentation retains more order information than shuffling after, and that even when shuffling after segmentation, position embeddings are meaningful because of the dependence between subword probability and sentence length.}
    \label{fig:motivating_plots}
\end{figure*}
Position embeddings are the only 
source of order information in these models; in their absence, contextual representations generated for tokens are independent of the actual position of the tokens in a sentence, and the models thus resemble heavily overparameterised bags-of-words. \citet{sinha2021masked} pre-trained RoBERTa models on shuffled corpora to demonstrate that the performance gap between these `shuffled' language models and models trained on unshuffled corpora is minor (when fine-tuned and evaluated downstream on the GLUE~\citep{WangGLUEMultiTaskBenchmark18} benchmark). They further show that this gap is considerably wider when a model is pre-trained without position embeddings. In this paper, we attempt to shed some light on why these models behave the way they do, and in doing so, seek to answer a set of pertinent questions: 
\begin{itemize}
    \item Do shuffled language models still have traces of word order information?
    \item Why is there a gap in performance between models {\em without} position embeddings and models trained on shuffled tokens, with the latter performing better? 
    \item Are there NLU benchmarks, other than GLUE, on which shuffled language models perform poorly?
\end{itemize}

\paragraph{Contributions} We first demonstrate, in Section~\ref{sec:order}, that shuffled language models {\em do} contain word order information, and are quite responsive to simple tests for word order information, particularly when compared to models trained without position representations. In Section~\ref{sec:mystery}, we demonstrate that pre-training is sufficient to learn this: position embeddings provide the appropriate inductive bias, and performing BPE segmentation after shuffling results in sensible n-grams appearing in the pre-training corpus; this gives models the capacity to learn word order within smaller local windows. Other minor cues - like correlations between sentence lengths and token distributions - also play a role. We further corroborate our analysis by examining attention patterns across models in Sec.~\ref{sec:attention}. %
In Section~\ref{sec:tasks}, we show that, while shuffled models might be almost as good as their un-shuffled counterparts on GLUE tasks, there exist NLU benchmarks that do require word order information to an extent that cannot be learned through fine-tuning alone. Finally, in Section~\ref{sec:other}, we describe miscellaneous experiments addressing the utility of positional embeddings when added just prior to fine-tuning.

\section{Models}
\citet{sinha2021masked} train several full-scale RoBERTa language models on the Toronto Book Corpus~\citep{Zhu_2015_ICCV} and English Wikipedia.\footnote{Training reportedly takes 72 hours on 64 GPUs.} Four of their models are trained on shuffled text, i.e., sentences in which $n$-grams are reordered at random.\footnote{The shuffling procedure does not reorder tokens {\em completely} at random, but moves a token in position $i$ to a {\em new} position selected at random among positions $j\neq i$.} We dub the original, unperturbed model \orig{}, and the scrambled models \sn{}, \snn{}, \snnn{} and \snnnn{} depending on the size of the shuffled $n$-grams: \sn{} reorders the unigrams in a sentence,  \snn{} reorders its bigrams, etc. For comparison, \citet{sinha2021masked} also train a RoBERTa language model entirely {\em without} position embeddings (\nopos{}), as well as a RoBERTa language model trained on a corpus drawn solely from unigram distributions of the original Book Corpus, i.e., a reshuffling of the entire corpus (\scorp{}). We experiment with their models, as well as with smaller models that we can train with a smaller carbon footprint. To this end, we downscale the RoBERTa architecture used in \citet{sinha2021masked}. Concretely, we train single-headed RoBERTa models, dividing the embedding and feed-forward dimensionality by 12, for 24 hours on a single GPU, on 100k sentences sampled from the Toronto Book Corpus. To this end, we train a custom vocabulary of size 5,000, which we use for indexing in all our subsequent experiments. While these smaller models are in no way meant to be fine-tuned and used downstream, they are useful proofs-of-concept that we later analyse.

\section{Probing for word order}
We begin by attempting to ascertain the extent to which shuffled language models are actually capable of encoding information pertaining to the naturalistic word order of sentences. We perform two simple tests on the full-scale models, in line with~\citet{wang2020position}: the first of these is a classification task where a logistic regressor is trained to predict whether a randomly sampled token precedes another in an unshuffled sentence, and the second involves predicting the position of a word in an unshuffled sentence. The fact that we \emph{do not} fine-tune any of the model parameters is noteworthy: the linear models can only learn word order information if it reflects in the representations the models generate somehow.

\label{sec:order}

\paragraph{Pairwise Classification}
For this experiment, we train a logistic regression classification model on word representations extracted from the final layer of the Transformer encoder, mean pooling over sub-tokens when required. For each word pair $x$ and $y$, the classifier is given a concatenation of our model $m$'s induced representations $m(x) \oplus m(y)$ and trained to predict a label indicating whether $x$ precedes $y$ or not. Holding out two randomly sampled positions, we use a training sets sized $2$k, $5$k, and $10$k, from the Universal Dependencies English-GUM corpus \citep{zeldes2017gum} (excluding sentences with more than 30 tokens to increase learnability) and a test set of size $2,000$. We report the mean accuracy from three runs. 

\paragraph{Regression}
Using the same data, we also train a ridge-regularised linear regression model to predict the position of a word $p(x)$ in an unshuffled sentence, given that word's model-induced representation $m(x)$. $R^2$ score is reported per model. To prevent the regressors from memorising word to position mappings, we perform 6-fold cross-validation, where the heldout part of the data contains no vocabulary overlap with the corresponding train set.

\paragraph{Results} For both tasks (see Table \ref{tab:classif_reg}), our results indicate that position encodings are particularly important for encoding word order: Classifiers and regressors trained on representations from \orig{} and \sn{} achieve high accuracies and $R^2$ scores, while those for \nopos{} are close to random. Both \orig{} and \sn{} appear to be better than random given only $2$k examples. These results imply that, given positional encodings and a modest training set of $2$k or more examples, a simple linear model is capable of extracting word order information, enabling almost perfect extrapolation to unseen positions. Whether the position encodings come from a model trained on natural or shuffled text does not appear to matter, emphasizing that shuffled language models do indeed contain substantial information about the original word order.


\begin{table}[t!]
    \centering
    \resizebox{\columnwidth}{!}{
    \begin{tabular}{l|ccc|c}
        \toprule
        \multirow{2}{*}{Model} & \multicolumn{3}{c|}{Classification (acc.)} & Regression ($R^2$) \\
        & 2k & 5k & 10k & -  \\
        \midrule
        \orig{} & 81.50 & 81.74 & 80.40 & 0.68 \\
        \sn{} & 65.96 & 64.98 & 71.82  &  0.60 \\
        \midrule 
        \nopos{} & 50.41 & 53.35 & 50.22 & 0.03 \\
        \bottomrule
    \end{tabular}}
\caption{Pairwise classification and regression results.}
\label{tab:classif_reg}
\end{table}

\section{Hidden word-order signals}\label{sec:mystery} 

In Section \ref{sec:order}, we observed that \newcite{sinha2021masked}'s shuffled language models surprisingly exhibit information about naturalistic word order. That these models contain positional information can also be seen by visualizing position embedding similarity. Figure~\ref{fig:pos_dot} displays Pearson correlations\footnote{We see similar patterns with dot products for all our plots; we use Pearson correlations to constrain our range to $[-1, 1]$.} for position embeddings with themselves, across positions. Here, we see that the shuffled models satisfy the idealised criteria for position embeddings described by~\citet{wang_positionembeddings_2021}: namely, they appear to be a) monotonous within smaller context windows, and b) invariant to translation. If position embedding correlations are consistent across offsets over the entire space of embeddings, the model can be said to have `learned' distances between tokens. Since transformers process all positions in parallel, and since language models without position embeddings do not exhibit such information, position embeddings have to be the source of this information. In what follows, we discuss this apparent paradox.  

\paragraph{Subword vs. word shuffling}
An important detail when running experiments on shuffled text, is {\em when} the shuffling operation takes place. When tokens are shuffled \emph{before} BPE segmentation, this leads to word-level shuffling, in which sequences of subwords that form words remain contiguous. Such sequences become a consistent, meaningful signal for language modelling, allowing models to efficiently utilise the inductive bias provided by position embeddings. Thus, even though our pretrained models have, in theory, not seen consecutive tokens in their pre-training data, they have learned to utilise positional embeddings to pay attention to adjacent tokens. The influence of this is somewhat visible in Figure~\ref{fig:motivating_plots}: while models trained on text shuffled before and after segmentation both exhibit shifts in the \emph{polarity} of their position correlations, only the former show bands of varying \emph{magnitude}, similar to the full-scale models. \citet{ravishankar-sogaard-2021-impact} discuss the implications of these patterns in a multilingual context; we hypothesise that in our context, the periodicity in magnitude is a visible artefact of the model's ability to leverage position embeddings to enable offset attention.
In Section~\ref{sec:attention}, we analyse the effect of shuffling the pre-training data on the models' attention mechanisms.

\begin{figure}
    \centering
    \includegraphics[width=0.49\textwidth]{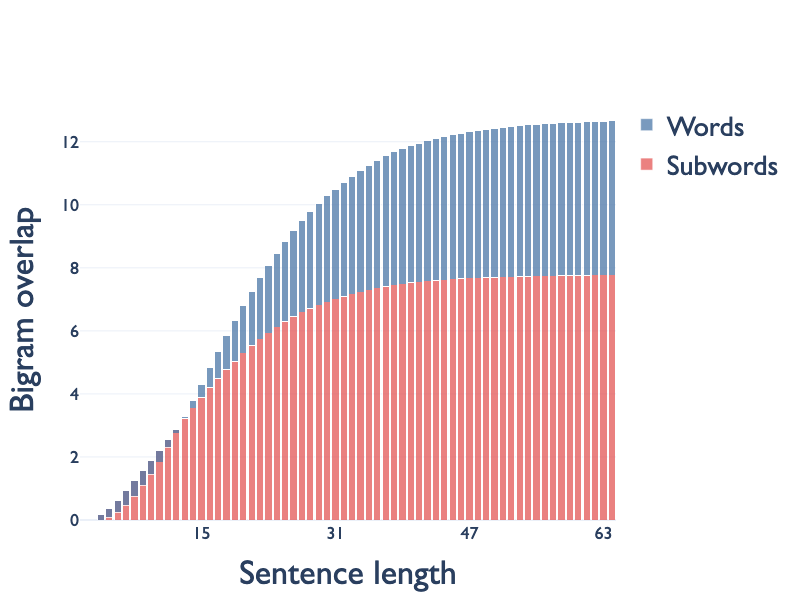}
    \caption{(Cumulative) plot showing subword bigram overlap after shuffling either words or subwords, as a percentage of the total number of seen bigrams. We see the overlap is significant, especially when performing shuffling before segmentation.}
    \label{fig:overlap}
\end{figure}

\paragraph{Accidental overlap}
In addition to the $n$-gram information which results from shuffling before segmentation, we also note that short sentences tend to include original bigrams with high probability, leading to stronger associations for words that are adjacent in the original texts. This effect is obviously much stronger when shuffling before segmentation than after segmentation. 
Figure~\ref{fig:overlap} shows how frequent overlapping bigrams (of any sort) are, comparing word and subword shuffling over 50k sentences.

\paragraph{Sentence length}
Finally, we observe some preserved information about the original word order even when shuffling is performed {\em after} segmentation. We hypothesize that this is a side-effect of the non-random relationship between sentence length and unigram probabilities. That unigram probabilities correlate with sentence length follows from the fact that different genres exhibit different sentence length distributions \cite{https://doi.org/10.1111/j.0039-3193.2004.00109.x,huiyan2017how}. Also, some words occur very frequently in formulaic contexts, e.g., {\em thank} in {\em thank you}. This potentially means that there is an approximately learnable relationship between the distribution of words and sentence boundary symbols. 

\begin{figure}
    \centering
    \includegraphics[width=0.49\textwidth]{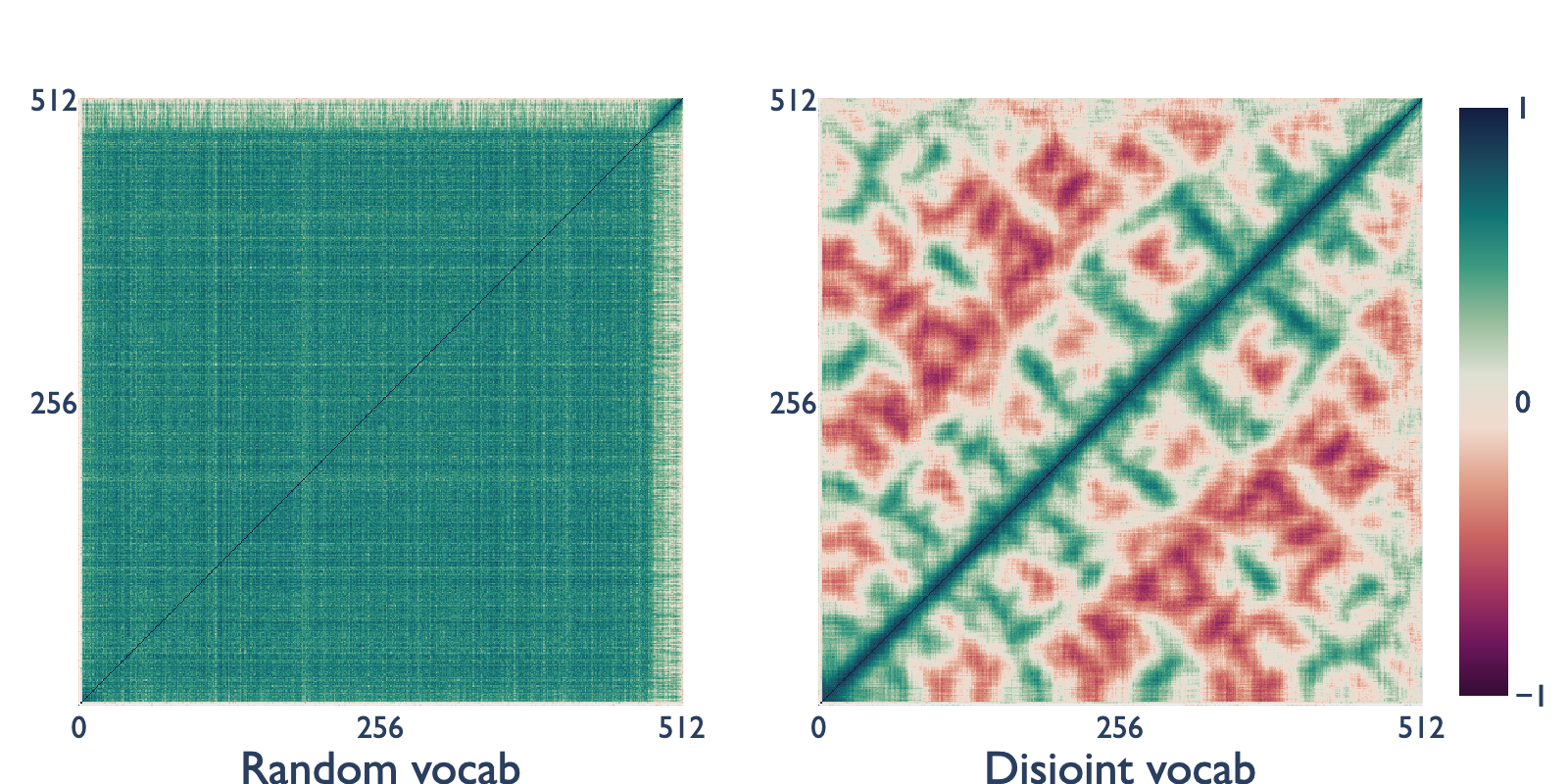}
    \caption{Similarity matrix between models with sentences sampled based on unigram corpus statistics; disjoint vocab implies a correlation between token choice and sentence length.}
    \label{fig:disjoint}
\end{figure}

To test for this, we train two smaller language models on unigram-sampled corpora: for the first, we use the first 100k BookCorpus sentences as our corpus, shuffling tokens at a corpus level (yet keeping the original sentence lengths). The stark difference in position embedding correlations between that and shuffling is seen in Figure~\ref{fig:motivating_plots}. For the second, we sample from two different unigram distributions: one for short sentences and one for longer sentences (details in Appendix~\ref{app:sampling}). While the first model induces no correlations at all, the second does, as shown in Figure~\ref{fig:disjoint}, implying that sentence length and unigram occurrences is enough to learn {\em some} order information. 

\section{Attention analysis}\label{sec:attention}
\begin{figure}
    \centering
    \includegraphics[width=0.49\textwidth]{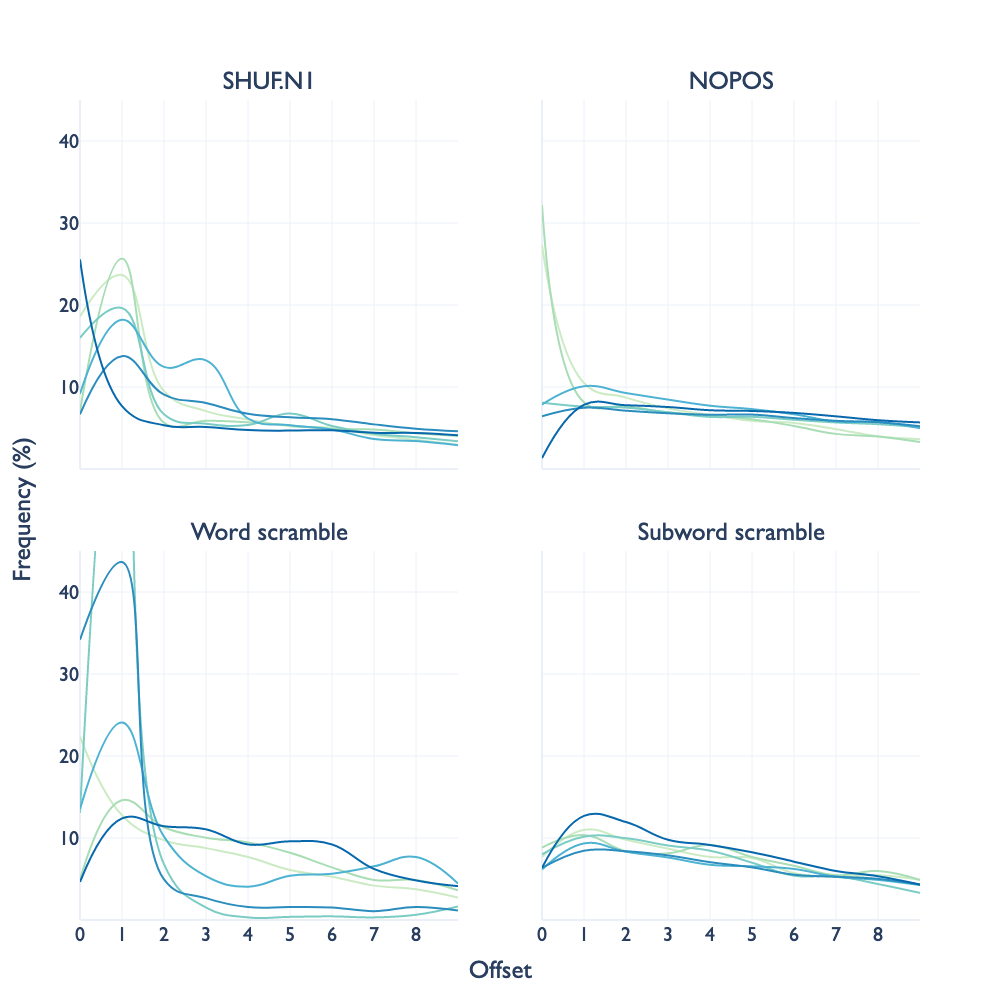}
    \caption{Relative frequency of offsets between token pairs in an attention relation; the y-axis denotes the percentage of total attention relations that occur at the offset indicated on the x-axis. We plot layers $l \in \{1, 2, 7, 8, 11, 12\}$ with increasing line darkness.}
    \label{fig:offset}
\end{figure}

Transformer-based language models commonly have attention heads that attend to neighboring positions \cite{voita-etal-2019-analyzing,ravishankar-etal-2021-attention}. Such attention heads are positional and can only be learned in the presence of order information. 
We attempt to visualise the attention mechanism for pre-trained models by calculating, for each head and layer, the offset between a token and the token that it pays maximum attention to\footnote{
This method of visualisation is somewhat limited, in that it examines only the \emph{maximum} attention paid by each token. We provide more detailed plots over attention \emph{distributions} in the Appendix.}. We then plot how frequent each offset is, as a percentage, over 100 Book Corpus sentences, in Figure~\ref{fig:offset}, where we present results for two full-scale models, and two smaller models (see \S2). When compared to \nopos{}, \sn{} has a less uniform pattern to its attention mechanism: it is likely, even at layer 0, to prefer to pay attention to adjacent tokens, somewhat mimicking a convolutional window~\citep{cordonnier_relationshipselfattention_2020a}. We see very similar differences in distribution between our smaller models: Shuffling after segmentation, i.e., at the subword level, influences early attention patterns.

\section{Evaluation beyond GLUE}\label{sec:tasks}
\paragraph{SuperGLUE and WinoGrande}
\newcite{sinha2021masked}'s investigation is conducted on GLUE and on the Paraphrase Adversaries
from Word shuffling (PAWS) dataset \cite{zhang2019paws}. For these datasets, they find that models pretrained on shuffled text perform only marginally worse than those pretrained on normal text. This result, they argue can be explained in two ways: either a) these tasks do not need word order information to be solved, or b) the required word order information can be acquired during finetuning. 
While GLUE has been a useful benchmark, several of the tasks which constitute it have been shown to be solvable using various spurious artefacts and heuristics \cite{gururangan-etal-2018-annotation, poliak-etal-2018-hypothesis}. If, for instance, through finetuning, models are learning to rely on such heuristics as lexical overlap for MNLI \cite{mccoy2019right}, then it is unsurprising that their performance is not greatly impacted by the lack of word order information. 

Evaluating on the more rigorous set of SuperGLUE tasks\footnote{Results are reported for an average of 3 runs per task.  The RTE task is excluded from our results as it is also part of GLUE; RTE results can be found in \newcite{sinha2021masked}.} \cite{wang2019superglue} and on the adversarially-filtered Winograd Schema examples \cite{levesque2012winograd} of the WinoGrande dataset \cite{sakaguchi2020winogrande} produces results which paint a more nuanced picture compared to those of \newcite{sinha2021masked}. The results, presented in Table \ref{tab:super_glue}, show accuracy or F1 scores for all models. For two of the tasks (MultiRC \cite{MultiRC2018}, COPA \cite{roemmele2011choice}), we observe a pattern in line with that seen in \newcite{sinha2021masked}'s GLUE and PAWS results: the drop in performance from \orig{} to \sn{} is minimal (mean: 1.75 points; mean across GLUE tasks: 3.3 points)\footnote{CoLA results are excluded from the GLUE calculations due to the very high variance across random seeds reported by \newcite{sinha2021masked}.}, while that to \nopos{} is more substantial (mean: 10.5 points; mean across GLUE tasks: 18.6 points).

This pattern alters for the BoolQ Yes/No question answering dataset \cite{clark2019boolq}, the CommitmentBank  \cite{de2019commitmentbank}, the ReCoRD reading comprehension dataset \cite{zhang2018record}, both the Winograd Schema tasks, and to some extent the Words in Context dataset  \cite{pilehvar2018wic}. For these tasks we observe a larger gap between  \orig{} and \sn{} (mean: 8.1 points), and an even larger one between  \orig{} and \nopos{} (mean: 19.78 points). 
We note that this latter set of tasks requires inferences which are more context-sensitive, in comparison to the two other tasks or to the GLUE tasks. 

Consider the Winograd schema tasks, for example. Each instance takes the form of a binary test with a statement comprising of two possible referents (blue) and a pronoun (red) such as: \texttt{\textcolor{blue}{Sid} explained his theory to \textcolor{blue}{Mark} but \textcolor{red}{he} couldn't \underline{convince} him.} The correct referent of the pronoun must be inferred based on a special discriminatory segment (underlined). In the above example, this depends on a) the identification of ``Sid'' as the subject of ``explained'' and b) inferring that the pronoun serving as the subject of ``convinced'' should refer to the same entity. Since the Winograd schema examples are designed so that the referents are equally associated with their context\footnote{e.g. Sid and Mark are both equally likely subjects/objects here. Not all Winograd schema examples are perfect in this regard, however, which could explain why scrambled models still perform above random. See \newcite{trichelair2018evaluation} for a discussion of the latter point.}, word order is crucial\footnote{Particularly in a language with limited morphological role marking such as English.} for establishing the roles of  ``Sid'' and ``Mark'' as subject and object of ``explained'' and ``he'' and ``him'' as those of ``convinced''. If these roles cannot be established, making the correct inference becomes impossible. 

A similar reasoning can be applied to the Words in Context dataset and the CommitmentBank. The former task tests the ability of a model to distinguish the senses of a polysemous word based on context. While this might often be feasible via a notion of contextual association that higher-order distributional statistics are sufficient for, some instances will require awareness of the word's role as an argument in the sentence. The latter task investigates the projectivity of finite clausal complements under entailment cancelling operators. This is dependent on both the scope of the entailment operator and the identity of the subject of the matrix predicate \cite{de2019commitmentbank}, both of which are sensitive to word order information.

A final consideration to take into account is dataset filtering. Two of the tasks where we observe the largest difference between \orig{}, \sn{}, and \nopos{} --- WinoGrande and ReCoRD --- apply filtering algorithms to remove cues or biases which would enable models to heuristically solve the tasks. This indicates that by filtering out examples containing cues that make them solvable via higher order statistics, such filtering strategies do succeed at compelling models to (at least partially) rely on word order information. 

\begin{table*}[t!]
{
    \centering
    \resizebox{\textwidth}{!}{
    \footnotesize
\begin{tabular}{c|ccccccccc}
        \toprule
        Model & BoolQ & CB & COPA &  MultiRC &  ReCoRD &  WiC & WSC & WinoGrande  \\
        \midrule
        \orig{} & 77.6 & 88.2 / 87.4  & 61.6 & 67.8 / 21.9  & 73.5 / 72.8 & 67.4 & 73.5 & 62.9 \\
        \sn{} &  72.4  &  79.7 / 82.5  & 59.7 & 66.2 / 15.0 &  61.1 / 60.4 & 63.0 &  62.9 & 55.7 \\
        \snn{} & 73.1  & 86.6 / 85.5 &   60.3 &  64.8 / 16.1 & 63.1 / 62.4 & 63.0 & 65.3 & 57.6 \\
        \snnnn{} &  73.5 & 87.9 / 87.1 & 60.8 & 66.2 / 18.2 & 64.6 / 63.9 & 62.4 & 65.3 &  59.53\\
        \nopos{} & 66.0 & 63.5 / 75.0 & 55.6 & 52.8 / 3.8 & 23.8 / 23.5 & 55.4 & 63.09 & 52.73 \\
        \scorp{} & 66.7 & 65.6 / 73.8 & 56.1 & 52.6 / 6.4 & 31.0 / 30.3 & 57.3 & 65.14 & 51.68\\
        \bottomrule
\end{tabular}}
 \caption{SuperGLUE and WinoGrande results for all models. Scores displayed are: Avg. F1 / Accuracy for CB; F1a / Exact Match for MultiRC; F1 / Accuracy for ReCoRD			; accuracy for the remaining tasks.}
\label{tab:super_glue}
}
\end{table*}

\begin{figure}
    \centering
    \includegraphics[width=0.48\textwidth]{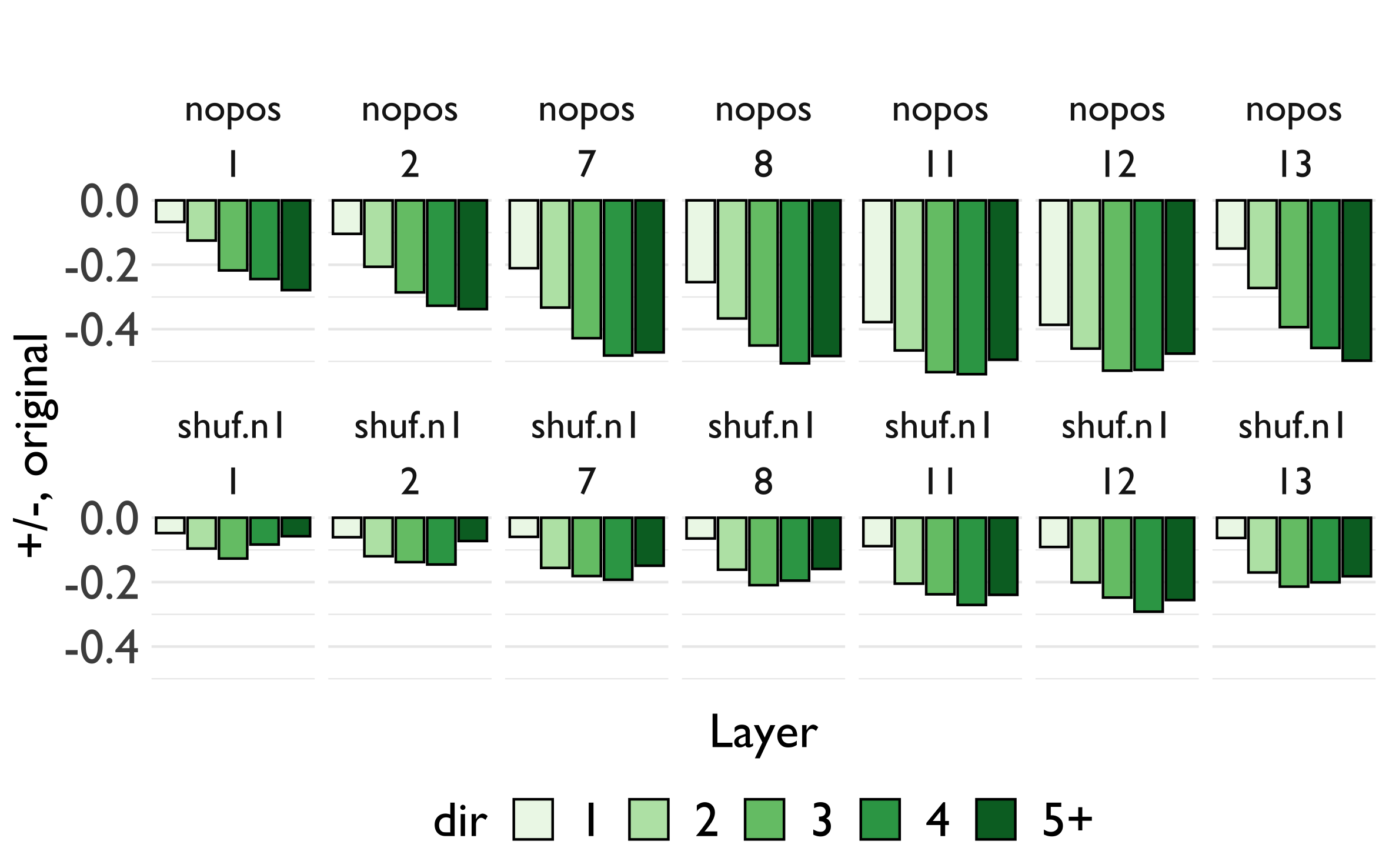}
    \caption{$\Delta$, dependency arcs probing accuracy across lengths 1-5+, w.r.t. \textsc{Orig}.}
    \label{fig:deplen}
\end{figure}

\paragraph{Dependency Tree Probing} Besides GLUE and PAWS, \citet{sinha2021masked}'s analysis also includes several probing experiments, wherein they attempt to decode dependency tree structure from model representations. They show, interestingly, that the \textsc{Shuf.N4, Shuf.N3} and \textsc{Shuf.N2} models perform only marginally worse than \textsc{Orig}, with \textsc{Shuf.N1} producing the lowest scores (lower, in fact, than \textsc{Shuf.Corpus}). Given the findings of Section \ref{sec:order}, we are interested in taking a closer look at this phenomenon. Here, we surmise that \textit{dependency length} plays a crucial role in the probing setup, where permuted models may succeed on par with \textsc{Orig} in capturing local, adjacent dependencies, but increasingly struggle to decode longer ones. To evaluate the extent to which this is true, we train a bilinear probe (used in \citet{hewitt-liang-2019-designing}) on top of all model representations and evaluate its accuracy across dependencies binned by length, where length between words $w_i$ and $w_j$ is defined as $|i-j|$. We opt for using the bilinear probe over the Pareto probing framework \citep{pimentel-etal-2020-pareto}, as the former learns a transformation directly over model representations, while the latter adds the parent and child MLP units from \citet{dozat-etal-2017-stanfords} -- acting more like a parser. We train probes on the English Web Treebank \citep{silveira2014gold} and evaluate using UAS, the standard parsing metric. 

Figure \ref{fig:deplen} shows $\Delta$ probing accuracy across various dependency lengths for \textsc{Nopos} and \textsc{Shuf.N1}, with respect to \textsc{Orig}\footnote{Note that Layer 13 refers to a linear mix of all model layers, as is done for ELMo \citep{peters-etal-2018-deep}.}; we include detailed $\Delta$s for all models in Appendix~\ref{app:delta}. For \nopos{}, parsing difficulty increases almost linearly with distance, often mimicking the actual frequency distribution of dependencies at these distances in the original treebank (Appendix~\ref{app:delta}); for \sn{}, the picture is a lot more nuanced, with dependencies at a distance of 1 consistently being closer in terms of parseability to \orig{}, which, we hypothesise, is due to its adjacency bias.



\section{Other Findings}\label{sec:other}

\paragraph{Random position embeddings are difficult to add post-training}
We tried to quantify the degree to which the inductive bias imparted by positional embeddings can be utilised, solely via fine-tuning. To do so, for a subset of GLUE tasks (MNLI, QNLI, RTE, SST-2, CoLA), we evaluate \nopos{}, and a variant where we randomly initialised learnable position embeddings and add them to the model, with the rest of the model equivalent to \nopos{}. We see no improvement in results, except for MNLI, that we hypothesise stems from position embeddings acting as some sort of regularisation parameter. To test this, we repeat the above set of experiments, this time injecting Gaussian noise instead; this has been empirically shown to have a regularising effect on the network~\citep{6796505,CamutoExplicitRegularisationGaussian21}. Adding Gaussian noise led to a slight increase in score for just MNLI, backing up our regularisation hypothesis.

\paragraph{Models learn to expect specific embeddings}

Replacing the positional embeddings in \orig{} with fixed, sinusoidal embeddings before fine-tuning significantly hurts scores on the same subset of GLUE tasks, implying that the models expect embeddings that resemble the inductive bias imparted by random embeddings, and that fine-tuning tasks do not have sufficient data to overcome this. The addition of fixed, \emph{sinusoidal} to \nopos{} also does not improve model performance on a similar subset of tasks; this implies, given that sinusoidal embeddings are already meaningful, that model weights also need to learn to fit the embeddings they are given, and that they need a substantial amount of data to do so.

\section{On Word Order}
\label{sec:related}
\paragraph{In Humans}
It is generally accepted that a majority of languages have ``canonical'' or ``base' word orderings  \cite{comrie1989language} (e.g. Subject-Verb-Object in English, and Subject-Object-Verb in Hindi).
Linguists consider word order to be a \textit{coding property} --- mechanisms by which abstract, syntactic structure is encoded in the surface form of utterances. Beyond word order, other coding properties include, e.g. subject-verb agreement, morphological case marking, or function words such as adpositions. In English, word order is among the most prominent coding properties, playing a crucial role in the expression of the main verb's core arguments: subject and object. For more morphologically complex languages, on the other hand, (e.g. Finnish and Turkish), word order is primarily used to convey pragmatic information such as topicalisation or focus. In such cases, argument structure is often signalled via case-marking, where numerous orderings become possible (shift in topic or focus nonwithstanding). We refer the reader to \citet{kulmizev2021schr} for a broader discussion of these topics and their implications when studying syntax through language models. 

More generally, evidence for the saliency of word order in linguistic processing and comprehension
comes from a variety of studies using acceptability judgements, eye-tracking data, and neural response measurements \cite{bever1970cognitive,danks1971psychological,just1980theory, friederici2000auditory, friederici2001syntactic,bahlmann2007fmri, lerner2011topographic, pallier2011cortical, fedorenko2016neural, ding2016cortical}. Psycholinguistic research has, however, also highlighted the robustness of sentence processing mechanisms to a variety of perturbations, including those which violate word order restrictions \cite{ferreira2002good, gibson2013rational, traxler2014trends}. In recent work, \newcite{mollica2020composition} tested the hypothesis that composition is the core function of the brain's language-selective network and that it can take place even when grammatical word order constrains are violated. Their findings confirmed this, showing that stimuli with shuffled word order where local dependencies were preserved --- as is, roughly speaking, the case for many dependencies in the sentences \snnnn{} is trained on --- elicited a neural response in the language network that is comparable to that elicited by normal sentences. When interword dependencies were disrupted so combinable words were so far apart that composition among nearby words was highly unlikely  --- as in \sn{}, neural response fell to a level compared to unconnected word lists. 


\paragraph{In Machines}
Recently, many NLP researchers have attempted to investigate the role of word order information in language models. For example, \newcite{lin2019open} employ diagnostic classifiers and attention analyses to demonstrate that lower (but not higher) layers of BERT encode word order information. \newcite{papadimitriou2021deep} find that Multilingual BERT is sensitive to morphosyntactic alignment, where numerous languages (out of 24 total) rely on word order to mark subjecthood (English among them). \newcite{alleman2021syntactic} implement an input perturbation framework (n-gram shuffling, phrase swaps, etc.), and employ it towards testing the sensitivity of BERT's
representations to various types of structure in sentences. They report a sensitivity to larger constituent units of sentences in higher layers, which they deduce to be influenced by hierarchical phrase structure. \newcite{o2021context} examine the contribution of various contextual features to the ability of GPT-2 \citep{radford2019language} to predict upcoming tokens. Their findings show that several destructive manipulations, including in-sentence word shuffling, applied to mid- and long range contexts lead only to a modest increase in \textit{usable information} as defined according to the V-information framework of \newcite{xu2020theory}. 

Similarly, word order information has been found not to be essential for various NLU tasks and datasets. Early work showed that Natural Language Inference tasks are largely insensitive to permutations of word order \cite{parikh2016decompos, sinha2020unnatural}. \newcite{pham2020out} and \newcite{gupta2021bert} discuss this in greater detail, demonstrating that test-time word order perturbations applied to GLUE benchmark tasks have little impact on LM performance. Following up on this, \newcite{sinha2021masked}, which our work builds on, found that \textit{pretraining} on scrambled text appears to only marginally affect model performance. Most related to this study, \newcite{clouatre2021demystifying} introduce two metrics for gauging the local and global ordering of tokens in scrambled texts, observing that only the latter is altered by the perturbation functions found in prior literature. In experiments with GLUE, they find that local (sub-word) perturbations show a substantially stronger performance decay compared to global ones. 

In this work, we present an in-depth analysis of these results, showing that LMs trained on scrambled text can actually retain word information and that -- as for humans -- their sensitivity to word order is dependent on a variety of factors such as the nature of the task and the locality of perturbation. While performance on some ``understanding'' evaluation tasks is not strongly affected by word order scrambling, the effect on others such as the Winograd Schema is far more evident. 



\section{Conclusion}
Much discussion has resulted from recent work showing that scrambling text at different stages of testing or training does not drastically alter the performance of language models on NLU tasks. In this work, we presented analyses painting a more nuanced picture of such findings. Primarily, we demonstrate that, as far as altered pre-training is concerned, models still do retain a semblance of word order knowledge --- largely at the local level. We show that this knowledge stems from cues in the altered data, such as adjacent BPE symbols and correlations between sentence length and content. The order in which shuffling is performed --- before or after BPE tokenization --- is influential in models' acquisition of word order, which calls for caution in interpreting previous results. Finally, we show that there exist NLU tasks that are far more sensitive to sentence structure as expressed by word order. 

\section*{Acknowledgements}
We thank Stephanie Brandl, Desmond Elliott, Yova Kementchedjhieva, Douwe Kiela and Miryam de Lhoneux for their feedback and comments. We acknowledge the CSC-IT Centre for Science, Finland, for providing computational resources. Vinit worked on this paper while on a research visit to the University of Copenhagen. Mostafa and Anders are supported by a Google Focused Research Award.

\bibliography{anthology,acl2020}

\begin{thebibliography}{63}
\expandafter\ifx\csname natexlab\endcsname\relax\def\natexlab#1{#1}\fi

\bibitem[{Alleman et~al.(2021)Alleman, Mamou, Del~Rio, Tang, Kim, and
  Chung}]{alleman2021syntactic}
Matteo Alleman, Jonathan Mamou, Miguel~A Del~Rio, Hanlin Tang, Yoon Kim, and
  SueYeon Chung. 2021.
\newblock Syntactic perturbations reveal representational correlates of
  hierarchical phrase structure in pretrained language models.
\newblock \emph{arXiv preprint arXiv:2104.07578}.

\bibitem[{Bahlmann et~al.(2007)Bahlmann, Rodriguez-Fornells, Rotte, and
  M{\"u}nte}]{bahlmann2007fmri}
J{\"o}rg Bahlmann, Antoni Rodriguez-Fornells, Michael Rotte, and Thomas~F
  M{\"u}nte. 2007.
\newblock An fmri study of canonical and noncanonical word order in german.
\newblock \emph{Human brain mapping}, 28(10):940--949.

\bibitem[{Bever(1970)}]{bever1970cognitive}
Thomas~G Bever. 1970.
\newblock The cognitive basis for linguistic structures.
\newblock \emph{Cognition and the development of language}.

\bibitem[{Bishop(1995)}]{6796505}
Chris~M. Bishop. 1995.
\newblock \href {https://doi.org/10.1162/neco.1995.7.1.108} {Training with
  noise is equivalent to tikhonov regularization}.
\newblock \emph{Neural Computation}, 7(1):108--116.

\bibitem[{Camuto et~al.(2021)Camuto, Willetts, {\c S}im{\c s}ekli, Roberts, and
  Holmes}]{CamutoExplicitRegularisationGaussian21}
Alexander Camuto, Matthew Willetts, Umut {\c S}im{\c s}ekli, Stephen Roberts,
  and Chris Holmes. 2021.
\newblock Explicit {{Regularisation}} in {{Gaussian Noise Injections}}.
\newblock \emph{arXiv:2007.07368 [cs, stat]}.

\bibitem[{Clark et~al.(2019)Clark, Lee, Chang, Kwiatkowski, Collins, and
  Toutanova}]{clark2019boolq}
Christopher Clark, Kenton Lee, Ming-Wei Chang, Tom Kwiatkowski, Michael
  Collins, and Kristina Toutanova. 2019.
\newblock Boolq: Exploring the surprising difficulty of natural yes/no
  questions.
\newblock \emph{arXiv preprint arXiv:1905.10044}.

\bibitem[{Clouatre et~al.(2021)Clouatre, Parthasarathi, Zouaq, and
  Chandar}]{clouatre2021demystifying}
Louis Clouatre, Prasanna Parthasarathi, Amal Zouaq, and Sarath Chandar. 2021.
\newblock Demystifying neural language models' insensitivity to word-order.
\newblock \emph{arXiv preprint arXiv:2107.13955}.

\bibitem[{Comrie(1989)}]{comrie1989language}
Bernard Comrie. 1989.
\newblock \emph{Language universals and linguistic typology: Syntax and
  morphology}.
\newblock University of Chicago press.

\bibitem[{Cordonnier et~al.(2020)Cordonnier, Loukas, and
  Jaggi}]{cordonnier_relationshipselfattention_2020a}
Jean-Baptiste Cordonnier, Andreas Loukas, and Martin Jaggi. 2020.
\newblock On the {{Relationship}} between {{Self}}-{{Attention}} and
  {{Convolutional Layers}}.
\newblock \emph{arXiv:1911.03584 [cs, stat]}.

\bibitem[{Danks and Glucksberg(1971)}]{danks1971psychological}
Joseph~H Danks and Sam Glucksberg. 1971.
\newblock Psychological scaling of adjective orders.
\newblock \emph{Journal of Memory and Language}, 10(1):63.

\bibitem[{De~Marneffe et~al.(2019)De~Marneffe, Simons, and
  Tonhauser}]{de2019commitmentbank}
Marie-Catherine De~Marneffe, Mandy Simons, and Judith Tonhauser. 2019.
\newblock The commitmentbank: Investigating projection in naturally occurring
  discourse.
\newblock In \emph{proceedings of Sinn und Bedeutung}, volume~23, pages
  107--124.

\bibitem[{Devlin et~al.(2018)Devlin, Chang, Lee, and
  Toutanova}]{devlin2018bert}
Jacob Devlin, Ming-Wei Chang, Kenton Lee, and Kristina Toutanova. 2018.
\newblock Bert: Pre-training of deep bidirectional transformers for language
  understanding.
\newblock \emph{arXiv preprint arXiv:1810.04805}.

\bibitem[{Ding et~al.(2016)Ding, Melloni, Zhang, Tian, and
  Poeppel}]{ding2016cortical}
Nai Ding, Lucia Melloni, Hang Zhang, Xing Tian, and David Poeppel. 2016.
\newblock Cortical tracking of hierarchical linguistic structures in connected
  speech.
\newblock \emph{Nature neuroscience}, 19(1):158--164.

\bibitem[{Dozat et~al.(2017)Dozat, Qi, and Manning}]{dozat-etal-2017-stanfords}
Timothy Dozat, Peng Qi, and Christopher~D. Manning. 2017.
\newblock \href {https://doi.org/10.18653/v1/K17-3002} {{S}tanford{'}s
  graph-based neural dependency parser at the {C}o{NLL} 2017 shared task}.
\newblock In \emph{Proceedings of the {C}o{NLL} 2017 Shared Task: Multilingual
  Parsing from Raw Text to Universal Dependencies}, pages 20--30, Vancouver,
  Canada. Association for Computational Linguistics.

\bibitem[{Fedorenko et~al.(2016)Fedorenko, Scott, Brunner, Coon, Pritchett,
  Schalk, and Kanwisher}]{fedorenko2016neural}
Evelina Fedorenko, Terri~L Scott, Peter Brunner, William~G Coon, Brianna
  Pritchett, Gerwin Schalk, and Nancy Kanwisher. 2016.
\newblock Neural correlate of the construction of sentence meaning.
\newblock \emph{Proceedings of the National Academy of Sciences},
  113(41):E6256--E6262.

\bibitem[{Ferreira et~al.(2002)Ferreira, Bailey, and
  Ferraro}]{ferreira2002good}
Fernanda Ferreira, Karl~GD Bailey, and Vittoria Ferraro. 2002.
\newblock Good-enough representations in language comprehension.
\newblock \emph{Current directions in psychological science}, 11(1):11--15.

\bibitem[{Friederici et~al.(2001)Friederici, Mecklinger, Spencer, Steinhauer,
  and Donchin}]{friederici2001syntactic}
Angela~D Friederici, Axel Mecklinger, Kevin~M Spencer, Karsten Steinhauer, and
  Emanuel Donchin. 2001.
\newblock Syntactic parsing preferences and their on-line revisions: A
  spatio-temporal analysis of event-related brain potentials.
\newblock \emph{Cognitive Brain Research}, 11(2):305--323.

\bibitem[{Friederici et~al.(2000)Friederici, Meyer, and
  Von~Cramon}]{friederici2000auditory}
Angela~D Friederici, Martin Meyer, and D~Yves Von~Cramon. 2000.
\newblock Auditory language comprehension: an event-related fmri study on the
  processing of syntactic and lexical information.
\newblock \emph{Brain and language}, 74(2):289--300.

\bibitem[{Gibson et~al.(2013)Gibson, Bergen, and
  Piantadosi}]{gibson2013rational}
Edward Gibson, Leon Bergen, and Steven~T Piantadosi. 2013.
\newblock Rational integration of noisy evidence and prior semantic
  expectations in sentence interpretation.
\newblock \emph{Proceedings of the National Academy of Sciences},
  110(20):8051--8056.

\bibitem[{Gupta et~al.(2021)Gupta, Kvernadze, and Srikumar}]{gupta2021bert}
Ashim Gupta, Giorgi Kvernadze, and Vivek Srikumar. 2021.
\newblock Bert \& family eat word salad: Experiments with text understanding.
\newblock \emph{arXiv preprint arXiv:2101.03453}.

\bibitem[{Gururangan et~al.(2018)Gururangan, Swayamdipta, Levy, Schwartz,
  Bowman, and Smith}]{gururangan-etal-2018-annotation}
Suchin Gururangan, Swabha Swayamdipta, Omer Levy, Roy Schwartz, Samuel Bowman,
  and Noah~A. Smith. 2018.
\newblock \href {https://doi.org/10.18653/v1/N18-2017} {Annotation artifacts in
  natural language inference data}.
\newblock In \emph{Proceedings of the 2018 Conference of the North {A}merican
  Chapter of the Association for Computational Linguistics: Human Language
  Technologies, Volume 2 (Short Papers)}, pages 107--112, New Orleans,
  Louisiana. Association for Computational Linguistics.

\bibitem[{Hewitt and Liang(2019)}]{hewitt-liang-2019-designing}
John Hewitt and Percy Liang. 2019.
\newblock \href {https://doi.org/10.18653/v1/D19-1275} {Designing and
  interpreting probes with control tasks}.
\newblock In \emph{Proceedings of the 2019 Conference on Empirical Methods in
  Natural Language Processing and the 9th International Joint Conference on
  Natural Language Processing (EMNLP-IJCNLP)}, pages 2733--2743, Hong Kong,
  China. Association for Computational Linguistics.

\bibitem[{Jin and Liu(2017)}]{huiyan2017how}
Huiyuan Jin and Haitao Liu. 2017.
\newblock \href {https://doi.org/10.1515/psicl-2017-0008} {How will text size
  influence the length of its linguistic constituents?}
\newblock \emph{Poznan Studies in Contemporary Linguistics}, 53.

\bibitem[{Just and Carpenter(1980)}]{just1980theory}
Marcel~A Just and Patricia~A Carpenter. 1980.
\newblock A theory of reading: From eye fixations to comprehension.
\newblock \emph{Psychological review}, 87(4):329.

\bibitem[{Khashabi et~al.(2018)Khashabi, Chaturvedi, Roth, Upadhyay, and
  Roth}]{MultiRC2018}
Daniel Khashabi, Snigdha Chaturvedi, Michael Roth, Shyam Upadhyay, and Dan
  Roth. 2018.
\newblock Looking beyond the surface:a challenge set for reading comprehension
  over multiple sentences.
\newblock In \emph{NAACL}.

\bibitem[{Kulmizev and Nivre(2021)}]{kulmizev2021schr}
Artur Kulmizev and Joakim Nivre. 2021.
\newblock Schrödinger's {T}ree -- {O}n {S}yntax and {N}eural {L}anguage
  {M}odels.
\newblock \emph{arXiv preprint arXiv:2110.08887}.

\bibitem[{Lerner et~al.(2011)Lerner, Honey, Silbert, and
  Hasson}]{lerner2011topographic}
Yulia Lerner, Christopher~J Honey, Lauren~J Silbert, and Uri Hasson. 2011.
\newblock Topographic mapping of a hierarchy of temporal receptive windows
  using a narrated story.
\newblock \emph{Journal of Neuroscience}, 31(8):2906--2915.

\bibitem[{Levesque et~al.(2012)Levesque, Davis, and
  Morgenstern}]{levesque2012winograd}
Hector Levesque, Ernest Davis, and Leora Morgenstern. 2012.
\newblock The winograd schema challenge.
\newblock In \emph{Thirteenth International Conference on the Principles of
  Knowledge Representation and Reasoning}.

\bibitem[{Lin et~al.(2019)Lin, Tan, and Frank}]{lin2019open}
Yongjie Lin, Yi~Chern Tan, and Robert Frank. 2019.
\newblock Open sesame: Getting inside bert's linguistic knowledge.
\newblock \emph{arXiv preprint arXiv:1906.01698}.

\bibitem[{Liu et~al.(2019)Liu, Ott, Goyal, Du, Joshi, Chen, Levy, Lewis,
  Zettlemoyer, and Stoyanov}]{LiuRoBERTaRobustlyOptimized19a}
Yinhan Liu, Myle Ott, Naman Goyal, Jingfei Du, Mandar Joshi, Danqi Chen, Omer
  Levy, Mike Lewis, Luke Zettlemoyer, and Veselin Stoyanov. 2019.
\newblock {{RoBERTa}}: A {{Robustly Optimized BERT Pretraining Approach}}.
\newblock \emph{arXiv:1907.11692 [cs]}.

\bibitem[{McCoy et~al.(2019)McCoy, Pavlick, and Linzen}]{mccoy2019right}
Tom McCoy, Ellie Pavlick, and Tal Linzen. 2019.
\newblock \href {https://doi.org/10.18653/v1/P19-1334} {Right for the wrong
  reasons: Diagnosing syntactic heuristics in natural language inference}.
\newblock In \emph{Proceedings of the 57th Annual Meeting of the Association
  for Computational Linguistics}, pages 3428--3448, Florence, Italy.
  Association for Computational Linguistics.

\bibitem[{Mollica et~al.(2020)Mollica, Siegelman, Diachek, Piantadosi,
  Mineroff, Futrell, Kean, Qian, and Fedorenko}]{mollica2020composition}
Francis Mollica, Matthew Siegelman, Evgeniia Diachek, Steven~T Piantadosi,
  Zachary Mineroff, Richard Futrell, Hope Kean, Peng Qian, and Evelina
  Fedorenko. 2020.
\newblock Composition is the core driver of the language-selective network.
\newblock \emph{Neurobiology of Language}, 1(1):104--134.

\bibitem[{O'Connor and Andreas(2021)}]{o2021context}
Joe O'Connor and Jacob Andreas. 2021.
\newblock What context features can transformer language models use?
\newblock \emph{arXiv preprint arXiv:2106.08367}.

\bibitem[{Pallier et~al.(2011)Pallier, Devauchelle, and
  Dehaene}]{pallier2011cortical}
Christophe Pallier, Anne-Dominique Devauchelle, and Stanislas Dehaene. 2011.
\newblock Cortical representation of the constituent structure of sentences.
\newblock \emph{Proceedings of the National Academy of Sciences},
  108(6):2522--2527.

\bibitem[{Papadimitriou et~al.(2021)Papadimitriou, Chi, Futrell, and
  Mahowald}]{papadimitriou2021deep}
Isabel Papadimitriou, Ethan~A Chi, Richard Futrell, and Kyle Mahowald. 2021.
\newblock Deep subjecthood: Higher-order grammatical features in multilingual
  bert.
\newblock \emph{arXiv preprint arXiv:2101.11043}.

\bibitem[{Parikh et~al.(2016)Parikh, T{\"a}ckstr{\"o}m, Das, and
  Uszkoreit}]{parikh2016decompos}
Ankur~P Parikh, Oscar T{\"a}ckstr{\"o}m, Dipanjan Das, and Jakob Uszkoreit.
  2016.
\newblock A decomposable attention model for natural language inference.
\newblock \emph{arXiv preprint arXiv:1606.01933}.

\bibitem[{Peters et~al.(2018)Peters, Neumann, Iyyer, Gardner, Clark, Lee, and
  Zettlemoyer}]{peters-etal-2018-deep}
Matthew Peters, Mark Neumann, Mohit Iyyer, Matt Gardner, Christopher Clark,
  Kenton Lee, and Luke Zettlemoyer. 2018.
\newblock \href {https://doi.org/10.18653/v1/N18-1202} {Deep contextualized
  word representations}.
\newblock In \emph{Proceedings of the 2018 Conference of the North {A}merican
  Chapter of the Association for Computational Linguistics: Human Language
  Technologies, Volume 1 (Long Papers)}, pages 2227--2237, New Orleans,
  Louisiana. Association for Computational Linguistics.

\bibitem[{Pham et~al.(2020)Pham, Bui, Mai, and Nguyen}]{pham2020out}
Thang~M Pham, Trung Bui, Long Mai, and Anh Nguyen. 2020.
\newblock Out of order: How important is the sequential order of words in a
  sentence in natural language understanding tasks?
\newblock \emph{arXiv preprint arXiv:2012.15180}.

\bibitem[{Pilehvar and Camacho-Collados(2018)}]{pilehvar2018wic}
Mohammad~Taher Pilehvar and Jose Camacho-Collados. 2018.
\newblock Wic: the word-in-context dataset for evaluating context-sensitive
  meaning representations.
\newblock \emph{arXiv preprint arXiv:1808.09121}.

\bibitem[{Pimentel et~al.(2020)Pimentel, Saphra, Williams, and
  Cotterell}]{pimentel-etal-2020-pareto}
Tiago Pimentel, Naomi Saphra, Adina Williams, and Ryan Cotterell. 2020.
\newblock \href {https://doi.org/10.18653/v1/2020.emnlp-main.254} {{P}areto
  probing: {T}rading off accuracy for complexity}.
\newblock In \emph{Proceedings of the 2020 Conference on Empirical Methods in
  Natural Language Processing (EMNLP)}, pages 3138--3153, Online. Association
  for Computational Linguistics.

\bibitem[{Poliak et~al.(2018)Poliak, Naradowsky, Haldar, Rudinger, and
  Van~Durme}]{poliak-etal-2018-hypothesis}
Adam Poliak, Jason Naradowsky, Aparajita Haldar, Rachel Rudinger, and Benjamin
  Van~Durme. 2018.
\newblock \href {https://doi.org/10.18653/v1/S18-2023} {Hypothesis only
  baselines in natural language inference}.
\newblock In \emph{Proceedings of the Seventh Joint Conference on Lexical and
  Computational Semantics}, pages 180--191, New Orleans, Louisiana. Association
  for Computational Linguistics.

\bibitem[{Radford et~al.(2019)Radford, Wu, Child, Luan, Amodei, Sutskever
  et~al.}]{radford2019language}
Alec Radford, Jeffrey Wu, Rewon Child, David Luan, Dario Amodei, Ilya
  Sutskever, et~al. 2019.
\newblock Language models are unsupervised multitask learners.
\newblock \emph{OpenAI blog}, 1(8):9.

\bibitem[{Ravishankar et~al.(2021)Ravishankar, Kulmizev, Abdou, S{\o}gaard, and
  Nivre}]{ravishankar-etal-2021-attention}
Vinit Ravishankar, Artur Kulmizev, Mostafa Abdou, Anders S{\o}gaard, and Joakim
  Nivre. 2021.
\newblock \href {https://doi.org/10.18653/v1/2021.eacl-main.264} {Attention can
  reflect syntactic structure (if you let it)}.
\newblock In \emph{Proceedings of the 16th Conference of the European Chapter
  of the Association for Computational Linguistics: Main Volume}, pages
  3031--3045, Online. Association for Computational Linguistics.

\bibitem[{Ravishankar and S{\o}gaard(2021)}]{ravishankar-sogaard-2021-impact}
Vinit Ravishankar and Anders S{\o}gaard. 2021.
\newblock \href {https://aclanthology.org/2021.emnlp-main.59} {The impact of
  positional encodings on multilingual compression}.
\newblock In \emph{Proceedings of the 2021 Conference on Empirical Methods in
  Natural Language Processing}, pages 763--777, Online and Punta Cana,
  Dominican Republic. Association for Computational Linguistics.

\bibitem[{Roemmele et~al.(2011)Roemmele, Bejan, and
  Gordon}]{roemmele2011choice}
Melissa Roemmele, Cosmin~Adrian Bejan, and Andrew~S Gordon. 2011.
\newblock Choice of plausible alternatives: An evaluation of commonsense causal
  reasoning.
\newblock In \emph{2011 AAAI Spring Symposium Series}.

\bibitem[{Sakaguchi et~al.(2020)Sakaguchi, Le~Bras, Bhagavatula, and
  Choi}]{sakaguchi2020winogrande}
Keisuke Sakaguchi, Ronan Le~Bras, Chandra Bhagavatula, and Yejin Choi. 2020.
\newblock Winogrande: An adversarial winograd schema challenge at scale.
\newblock In \emph{Proceedings of the AAAI Conference on Artificial
  Intelligence}, volume~34, pages 8732--8740.

\bibitem[{Sigurd et~al.(2004)Sigurd, Eeg-Olofsson, and
  Van~Weijer}]{https://doi.org/10.1111/j.0039-3193.2004.00109.x}
Bengt Sigurd, Mats Eeg-Olofsson, and Joost Van~Weijer. 2004.
\newblock \href
  {https://doi.org/https://doi.org/10.1111/j.0039-3193.2004.00109.x} {Word
  length, sentence length and frequency – zipf revisited}.
\newblock \emph{Studia Linguistica}, 58(1):37--52.

\bibitem[{Silveira et~al.(2014)Silveira, Dozat, De~Marneffe, Bowman, Connor,
  Bauer, and Manning}]{silveira2014gold}
Natalia Silveira, Timothy Dozat, Marie-Catherine De~Marneffe, Samuel~R Bowman,
  Miriam Connor, John Bauer, and Christopher~D Manning. 2014.
\newblock A gold standard dependency corpus for english.
\newblock In \emph{LREC}, pages 2897--2904. Citeseer.

\bibitem[{Sinha et~al.(2021)Sinha, Jia, Hupkes, Pineau, Williams, and
  Kiela}]{sinha2021masked}
Koustuv Sinha, Robin Jia, Dieuwke Hupkes, Joelle Pineau, Adina Williams, and
  Douwe Kiela. 2021.
\newblock \href {https://doi.org/10.18653/v1/2021.emnlp-main.230} {Masked
  language modeling and the distributional hypothesis: Order word matters
  pre-training for little}.
\newblock In \emph{Proceedings of the 2021 Conference on Empirical Methods in
  Natural Language Processing}, pages 2888--2913, Online and Punta Cana,
  Dominican Republic. Association for Computational Linguistics.

\bibitem[{Sinha et~al.(2020)Sinha, Parthasarathi, Pineau, and
  Williams}]{sinha2020unnatural}
Koustuv Sinha, Prasanna Parthasarathi, Joelle Pineau, and Adina Williams. 2020.
\newblock Unnatural language inference.
\newblock \emph{arXiv preprint arXiv:2101.00010}.

\bibitem[{Traxler(2014)}]{traxler2014trends}
Matthew~J Traxler. 2014.
\newblock Trends in syntactic parsing: Anticipation, bayesian estimation, and
  good-enough parsing.
\newblock \emph{Trends in cognitive sciences}, 18(11):605--611.

\bibitem[{Trichelair et~al.(2018)Trichelair, Emami, Cheung, Trischler, Suleman,
  and Diaz}]{trichelair2018evaluation}
Paul Trichelair, Ali Emami, Jackie Chi~Kit Cheung, Adam Trischler, Kaheer
  Suleman, and Fernando Diaz. 2018.
\newblock On the evaluation of common-sense reasoning in natural language
  understanding.
\newblock \emph{arXiv preprint arXiv:1811.01778}.

\bibitem[{Vaswani et~al.(2017)Vaswani, Shazeer, Parmar, Uszkoreit, Jones,
  Gomez, Kaiser, and Polosukhin}]{vaswani2017attention}
Ashish Vaswani, Noam Shazeer, Niki Parmar, Jakob Uszkoreit, Llion Jones,
  Aidan~N Gomez, {\L}ukasz Kaiser, and Illia Polosukhin. 2017.
\newblock Attention is all you need.
\newblock In \emph{Advances in neural information processing systems}, pages
  5998--6008.

\bibitem[{Voita et~al.(2019)Voita, Talbot, Moiseev, Sennrich, and
  Titov}]{voita-etal-2019-analyzing}
Elena Voita, David Talbot, Fedor Moiseev, Rico Sennrich, and Ivan Titov. 2019.
\newblock \href {https://doi.org/10.18653/v1/P19-1580} {Analyzing multi-head
  self-attention: Specialized heads do the heavy lifting, the rest can be
  pruned}.
\newblock In \emph{Proceedings of the 57th Annual Meeting of the Association
  for Computational Linguistics}, pages 5797--5808, Florence, Italy.
  Association for Computational Linguistics.

\bibitem[{Wang et~al.(2019)Wang, Pruksachatkun, Nangia, Singh, Michael, Hill,
  Levy, and Bowman}]{wang2019superglue}
Alex Wang, Yada Pruksachatkun, Nikita Nangia, Amanpreet Singh, Julian Michael,
  Felix Hill, Omer Levy, and Samuel Bowman. 2019.
\newblock {{SuperGLUE}}: {{A}} stickier benchmark for general-purpose language
  understanding systems.
\newblock In \emph{Advances in Neural Information Processing Systems},
  volume~32. {Curran Associates, Inc.}

\bibitem[{Wang et~al.(2018)Wang, Singh, Michael, Hill, Levy, and
  Bowman}]{WangGLUEMultiTaskBenchmark18}
Alex Wang, Amanpreet Singh, Julian Michael, Felix Hill, Omer Levy, and Samuel
  Bowman. 2018.
\newblock \href {https://doi.org/10.18653/v1/W18-5446} {{{GLUE}}: A
  {{Multi}}-{{Task Benchmark}} and {{Analysis Platform}} for {{Natural Language
  Understanding}}}.
\newblock In \emph{Proceedings of the 2018 {{EMNLP Workshop BlackboxNLP}}:
  Analyzing and {{Interpreting Neural Networks}} for {{NLP}}}, pages 353--355,
  {Brussels, Belgium}. {Association for Computational Linguistics}.

\bibitem[{Wang et~al.(2021)Wang, Shang, Lioma, Jiang, Yang, Liu, and
  Simonsen}]{wang_positionembeddings_2021}
Benyou Wang, Lifeng Shang, Christina Lioma, Xin Jiang, Hao Yang, Qun Liu, and
  Jakob~Grue Simonsen. 2021.
\newblock {{ON POSITION EMBEDDINGS IN BERT}}.
\newblock page~21.

\bibitem[{Wang and Chen(2020)}]{wang2020position}
Yu-An Wang and Yun-Nung Chen. 2020.
\newblock \href {https://doi.org/10.18653/v1/2020.emnlp-main.555} {What do
  position embeddings learn? an empirical study of pre-trained language model
  positional encoding}.
\newblock In \emph{Proceedings of the 2020 Conference on Empirical Methods in
  Natural Language Processing (EMNLP)}, pages 6840--6849, Online. Association
  for Computational Linguistics.

\bibitem[{Xu et~al.(2020)Xu, Zhao, Song, Stewart, and Ermon}]{xu2020theory}
Yilun Xu, Shengjia Zhao, Jiaming Song, Russell Stewart, and Stefano Ermon.
  2020.
\newblock A theory of usable information under computational constraints.
\newblock \emph{arXiv preprint arXiv:2002.10689}.

\bibitem[{Zeldes(2017)}]{zeldes2017gum}
Amir Zeldes. 2017.
\newblock The gum corpus: Creating multilayer resources in the classroom.
\newblock \emph{Language Resources and Evaluation}, 51(3):581--612.

\bibitem[{Zhang et~al.(2018)Zhang, Liu, Liu, Gao, Duh, and
  Van~Durme}]{zhang2018record}
Sheng Zhang, Xiaodong Liu, Jingjing Liu, Jianfeng Gao, Kevin Duh, and Benjamin
  Van~Durme. 2018.
\newblock Record: Bridging the gap between human and machine commonsense
  reading comprehension.
\newblock \emph{arXiv preprint arXiv:1810.12885}.

\bibitem[{Zhang et~al.(2019)Zhang, Baldridge, and He}]{zhang2019paws}
Yuan Zhang, Jason Baldridge, and Luheng He. 2019.
\newblock \href {https://doi.org/10.18653/v1/N19-1131} {{PAWS}: Paraphrase
  adversaries from word scrambling}.
\newblock In \emph{Proceedings of the 2019 Conference of the North {A}merican
  Chapter of the Association for Computational Linguistics: Human Language
  Technologies, Volume 1 (Long and Short Papers)}, pages 1298--1308,
  Minneapolis, Minnesota. Association for Computational Linguistics.

\bibitem[{Zhu et~al.(2015)Zhu, Kiros, Zemel, Salakhutdinov, Urtasun, Torralba,
  and Fidler}]{Zhu_2015_ICCV}
Yukun Zhu, Ryan Kiros, Rich Zemel, Ruslan Salakhutdinov, Raquel Urtasun,
  Antonio Torralba, and Sanja Fidler. 2015.
\newblock Aligning books and movies: Towards story-like visual explanations by
  watching movies and reading books.
\newblock In \emph{The IEEE International Conference on Computer Vision
  (ICCV)}.

\end{thebibliography}
\bibliographystyle{acl_natbib}

\appendix
\section{Subword vs. word scrambling}
\label{app:plots}
\begin{figure*}[htp]
    \centering
    \includegraphics[width=\textwidth]{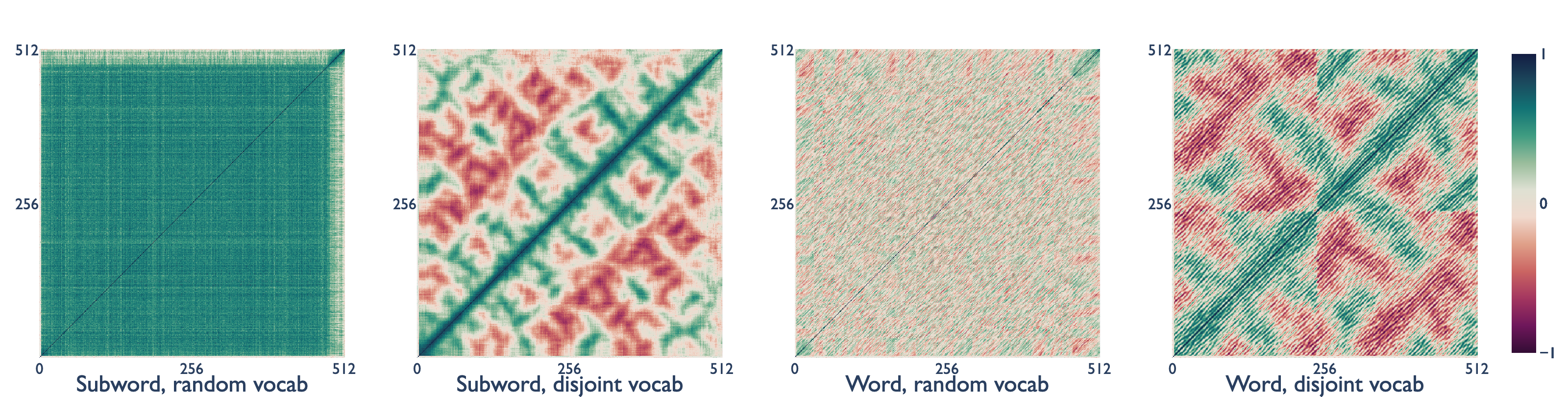}
    \caption{Pearson correlations, when scrambling by subword/word, with/without disjoint vocabularies. Disjoint vocabularies appear to induce patterns in position-position correlations, while scrambling at a word level induces `stripes' of oscillating magnitude; this is likely due to position embeddings learning connections to adjacent tokens. }
\end{figure*}

\section{On biased sampling}
\label{app:sampling}
We first split our vocab of size 5,000 into two halves, both of size 2500, such that the sum total of unigram frequencies of tokens in each half is roughly equivalent. Next, iterating over 100k BookCorpus sentences, we determine the sentence length $l$, for which there are an equivalent number of tokens in sentences with length $< l$ and sentences with length $>= l$. We then sample tokens from the first vocab half for sentences $< l$, and from the second vocab half for sentences with length $>= l$, 80\% of the time; for the other 20\%, we sample from the opposite half to introduce some overlap.

\section{Full UD results}
\label{app:delta}

\begin{figure}[h]
    \centering
    \includegraphics[width=0.49\textwidth,height=0.45\textwidth]{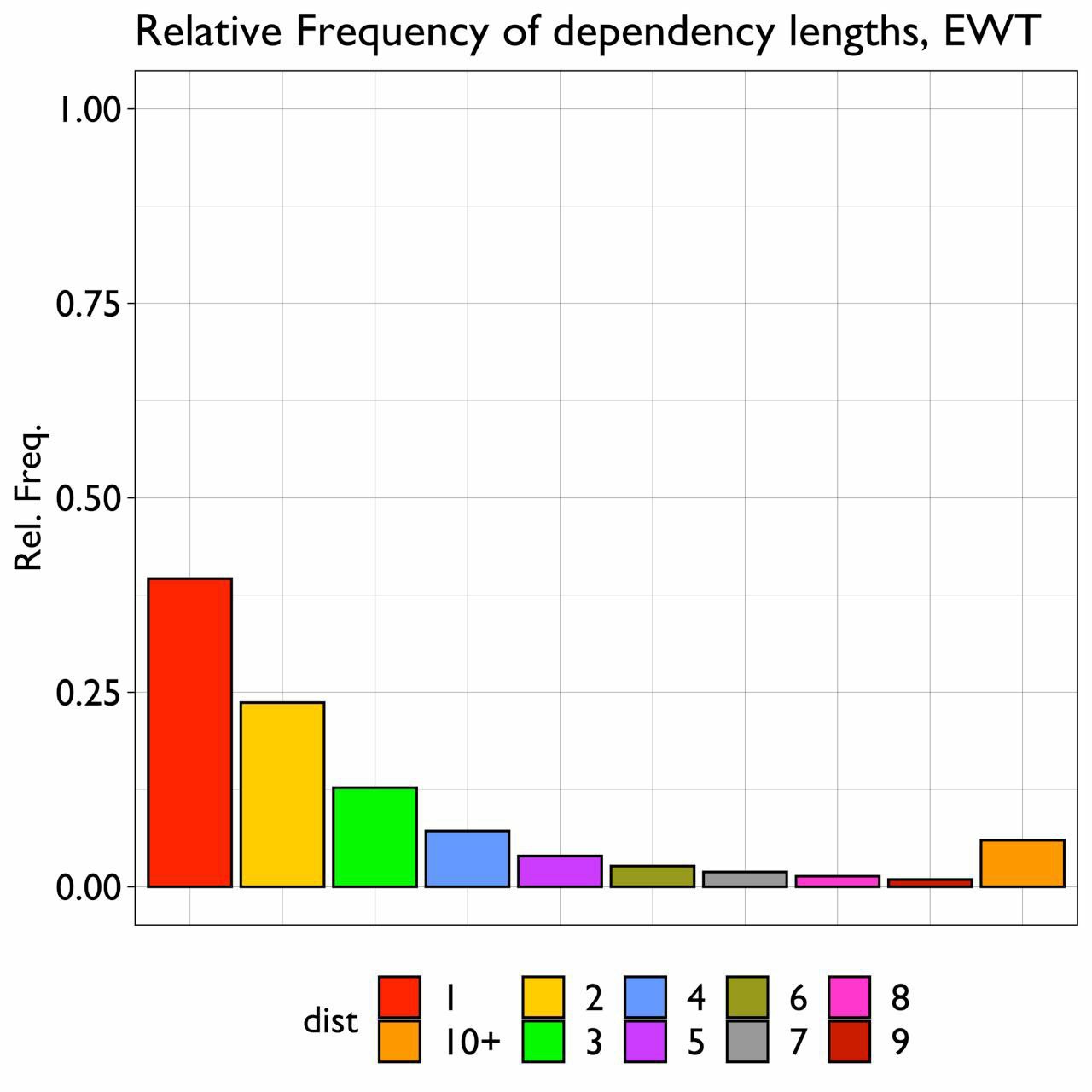}
    \caption{Relative frequencies of dependency relations in $UD_{English-EWT}$, at a dependency lengths indicated by the x-axis}
\end{figure}

\begin{figure}[h]
    \centering
    \includegraphics[width=0.49\textwidth]{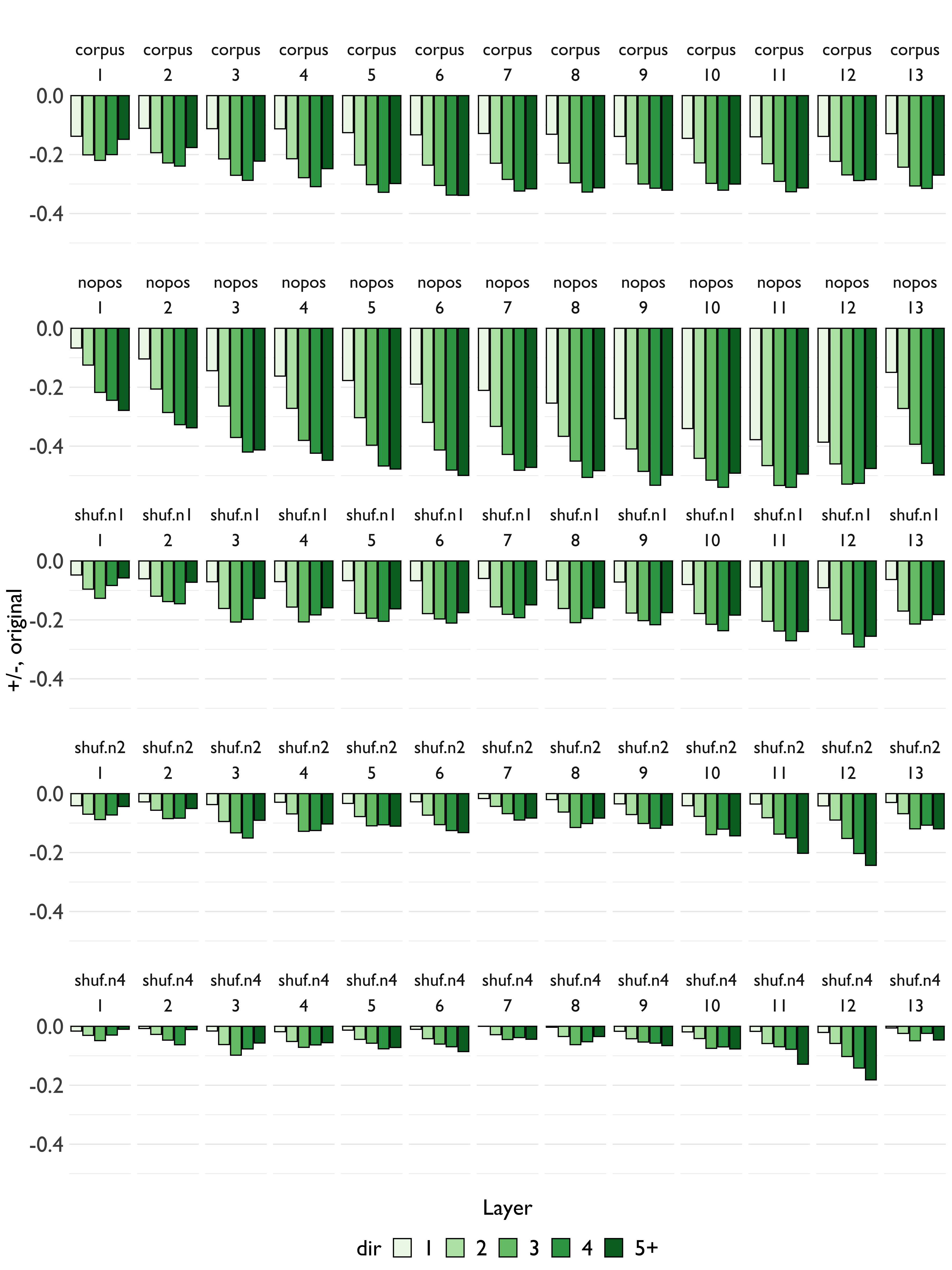}
    \caption{$\Delta$ UAS, all models and layers across dependency lengths 1-5+, w.r.t. \textsc{Orig}. Layer 13 represents a linear mix of all model layers. }
    \label{fig:deplen_all}
\end{figure}

\end{document}